%% file: emnlp2018.tex
\documentclass[11pt,a4paper]{article}
\usepackage[hyperref]{emnlp2018}
\usepackage{times}
\usepackage{latexsym}

\usepackage{microtype}
\usepackage{booktabs}
\usepackage{url}
\usepackage{multirow}
\usepackage{amsmath}
\usepackage{amsfonts}
\usepackage{graphicx}
\usepackage{enumitem} % Reduce space between list.
\usepackage[skip=3pt]{caption} % Reduce space before caption.
\aclfinalcopy % Uncomment this line for the final submission

\title{Automatic Event Salience Identification}

\author{Zhengzhong Liu \qquad Chenyan Xiong \qquad Teruko Mitamura \qquad Eduard Hovy \\
  Language Technologies Institute \\
  Carnegie Mellon University \\
  Pittsburgh, PA 15213, USA \\
  {\tt \{liu, cx, teruko, hovy\}@cs.cmu.edu}}

\date{}

\begin{document}
\maketitle
\begin{abstract}
Identifying the salience (i.e.\ importance) of discourse units is an important task in language understanding. While events play important roles in text documents, little research exists on analyzing their saliency status. This paper empirically studies the \emph{Event Salience} task and proposes two salience detection models based on content similarities and discourse relations. The first is a feature based salience model that incorporates similarities among discourse units. The second is a neural model that captures more complex relations between discourse units. 
Tested on our new large-scale event salience corpus,
both methods significantly outperform the strong frequency baseline, while our neural model further improves the feature based one by a large margin. 
Our analyses demonstrate that our neural model captures interesting connections between salience and discourse unit relations (e.g., scripts and frame structures). 
% The studies are all conducted on a new large-scale event salience corpus. 
\end{abstract}
\input{1.introduction}

\input{2.related_work}
\input{3.corpus}
\input{4.feature-based}
\input{5.neural-based}

\input{6.experiment}
\input{7.evaluation}
\input{8.conclusion}

\section*{Acknowledgement}

This research was supported by DARPA grant FA8750-18-2-0018 funded under the AIDA program and National Science Foundation (NSF) grant IIS-1422676. Any opinions, findings, and conclusions in this paper are the authors’ and do not necessarily reflect the sponsors’. We thank the anonymous reviewers whose suggestions helped clarify this paper.

\bibliography{emnlp2018}
\bibliographystyle{acl_natbib_nourl}

\end{document}

%% file: 1.introduction.tex
\section{Introduction}\label{sec:introduction}
Automatic extraction of prominent information from text has always been a core problem in language research. 
While traditional methods mostly concentrate on the word level, researchers start to analyze higher-level discourse units in text, such as entities~\cite{Dunietz2014} and events~\cite{Choubey2018}.

\begin{figure}
\frame{\includegraphics[width=1.0\linewidth]{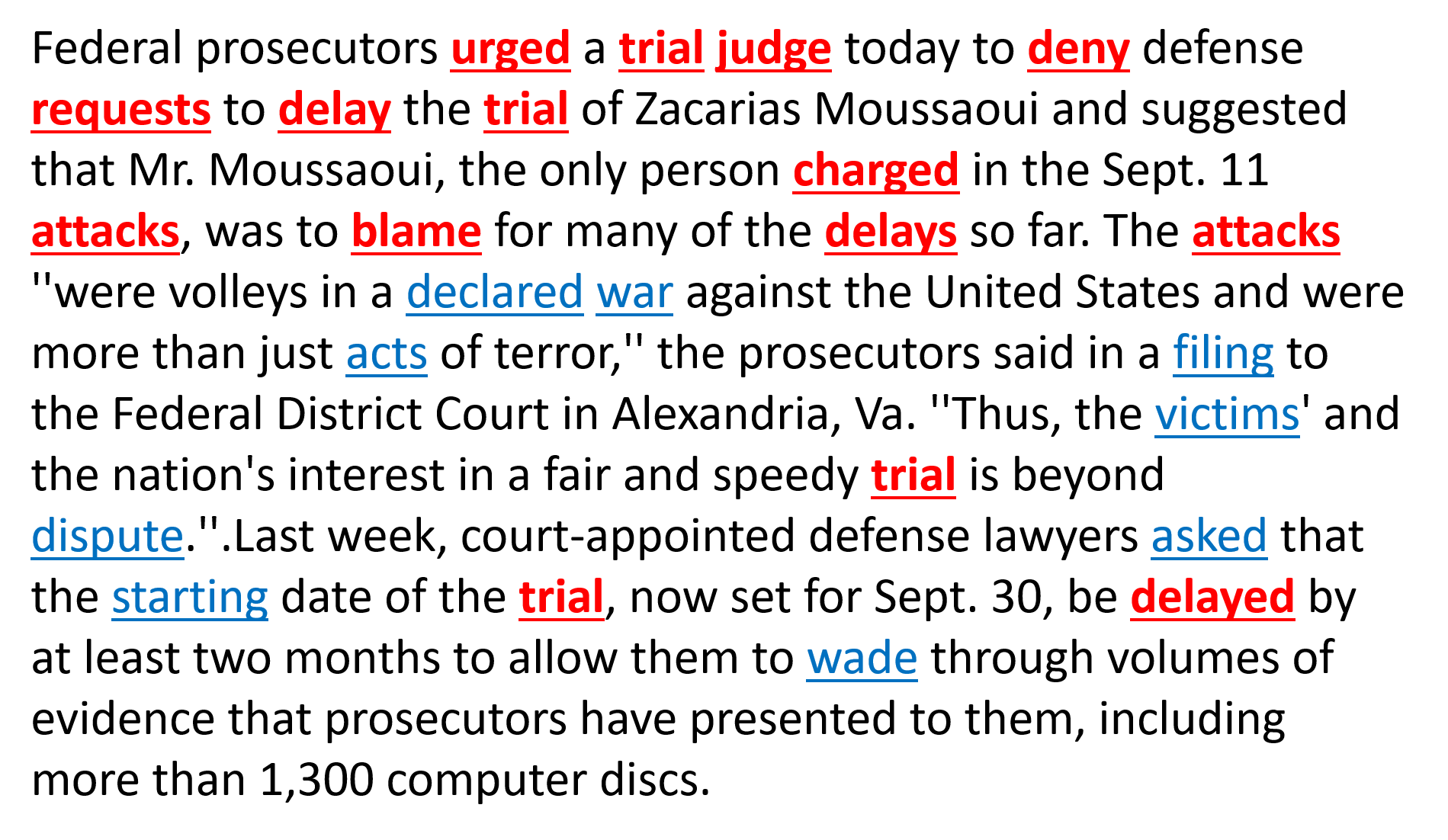}}
\caption{Examples annotations. Underlying words are annotated event triggers; the red bold ones are annotated as salient.\label{fig:annotation_example}
}
\end{figure}

%Application motivation
Events are important discourse units that form the backbone of our communication. They play various roles in documents. Some are more central in discourse: connecting other entities and events, or providing key information of a story. Others are less relevant, 
% creating burden for language understanding. 
% which may be noisy for NLP systems.
but not easily identifiable by NLP systems.
Hence it is important to be able to quantify the ``importance'' of events. For example, Figure~\ref{fig:annotation_example} is a news excerpt describing a debate around a jurisdiction process: ``\textit{trial}'' is central as the main discussing topic, while ``\textit{war}'' is not.

% Practical motivation
Researchers are aware of the need to identify central events in applications like detecting salient relations~\cite{Zhang2015}, and identifying climax in storyline~\cite{Vossen2015}. Generally, the salience of discourse units is important for language understanding tasks, such as document analysis~\cite{Barzilay2008}, information retrieval~\cite{Xiong2018}, and semantic role labeling~\cite{Cheng2018}. Thus, proper models for finding important events are desired.

% Theory motivation
In this work, we study the task of \textbf{event salience detection}, to find events that are most relevant to the main content of documents.
To build a salience detection model, one core observation is that salient discourse units are forming discourse relations. In Figure~\ref{fig:annotation_example}, the ``\textit{trial}'' event is connected to many other events: ``\textit{charge}'' is pressed before ``\textit{trial}''; ``\textit{trial}'' is being ``\textit{delayed}''. 

% Model design based on the motivations
We present two salience detection systems based on the observations. First is a feature based learning to rank model. Beyond basic features like frequency and discourse location, we design features using cosine similarities among events and entities, to estimate the \textit{content organization}~\cite{Grimes1975}: how lexical meaning of elements relates to each other. Similarities from within-sentence or across the whole document are used to capture interactions on both local and global aspects (\S\ref{sec:feature_model}). The model significantly outperforms a strong ``Frequency'' baseline in our experiments.

However, there are other discourse relations beyond lexical similarity. 
Figure~\ref{fig:annotation_example} showcases some: the script relation~\cite{Schank1977}\footnote{Scripts are prototypical sequences of events: a \textit{restaurant} script normally contains events like ``order'', ``eat'' and ``pay''.}  between ``\textit{charge}'' and ``\textit{trial}'', and the frame relation~\cite{Baker1998} between ``\textit{attacks}'' and ``\textit{trial}'' (``\textit{attacks}'' fills the ``charges'' role of ``\textit{trial}''). Since it is unclear which ones contribute more to salience, we design a Kernel based Centrality Estimation (\texttt{KCE}) model (\S\ref{sec:neural_model}) to capture salient specific interactions between discourse units automatically.

In \texttt{KCE}, discourse units are projected to embeddings, which are trained end-to-end towards the salience task to capture rich semantic information. A set of soft-count kernels are trained to weigh salient specific latent relations between discourse units. With the capacity to model richer relations, \texttt{KCE} outperforms the feature-based model by a large margin (\S\ref{sec:results}). Our analysis shows that \texttt{KCE} is exploiting several relations between discourse units: including script and frames (Table~\ref{tab:kernel_examples}). To further understand the nature of \texttt{KCE}, we conduct an \textit{intrusion test} (\S\ref{sec:extrinsic}), which requires a model to identify events from another document. The test shows salient events form tightly related groups with relations captured by \texttt{KCE}\@.
% We observe that salient events are likely to form self-cohesive groups.

The notion of salience is subjective and may vary from person to person. We follow the empirical approaches used in entity salience research~\cite{Dunietz2014}. We consider the \textit{summarization test}: an event is considered salient if a summary written by a human is likely to include it,
% . This is consistent with our definition
since events about the main content are more likely to appear in a summary. This approach allows us to create a large-scale corpus (\S\ref{sec:corpus}). 

% Contribution
In this paper, we make three main contributions. First, we present two event salience detection systems, which capture rich relations among discourse units. Second, we observe interesting connections between salience and various discourse relations (\S\ref{sec:results} and Table~\ref{tab:kernel_examples}), implying potential research on these areas. Finally, we construct a large scale event salience corpus, providing a testbed for future research. Our code, dataset and models are publicly available\footnote{\url{https://github.com/hunterhector/EventSalience}}.

%% file: 2.related_work.tex
\section{Related Work}

Events have been studied on many aspects due to their importance in language. To name a few: event detection~\cite{Li2013,Nguyen2015,Peng2016}, coreference~\cite{Liu2013,Lu2017}, temporal analysis~\cite{Do2012,Chambers2014}, sequencing~\cite{Araki2013}, script induction~\cite{Chambers2008,Balasubramanian2013,Rudinger2015,Pichotta2016}. 

However, studies on event salience are premature. Some previous work attempts to approximate event salience with word frequency or discourse position~\cite{Vossen2015,Zhang2015}. Parallel to ours, \newcite{Choubey2018} propose a task to find the most dominant event in news articles. They draw connections between event coreference and importance, on hundreds of closed-domain documents, using several oracle event attributes. In contrast, our proposed models are fully learned and applied on more general domains and at a larger scale. We also do not restrict to a single most important event per document.

There is a small but growing line of work on entity salience~\cite{Dunietz2014,Dojchinovski2016,Xiong2018,Ponza2018}. In this work, we study the case for events.

Text relations have been studied in tasks like text summarization, which mainly focused on cohesion~\cite{Halliday1976}. 
% The assumption is that important units are the most connected ones in discourse. 
Grammatical cohesion methods make use of document level structures such as anaphora relations~\cite{Baldwin1998} and discourse parse trees~\cite{Marcu1999}. Lexical cohesion based methods focus on repetitions and synonyms on the lexical level~\cite{Skorochodko.1971, Morris1991,Erkan2004}. Though sharing similar intuitions, our proposed models are designed to learn richer semantic relations in the embedding space. 

Comparing to the traditional summarization task, we focus on events, which are at a different granularity. Our experiments also unveil interesting phenomena among events and other discourse units.

%% file: 3.corpus.tex
\section{The Event Salience Corpus}\label{sec:corpus}
This section introduces our approach to construct a large-scale event salience corpus, including methods for finding event mentions and obtaining saliency labels. The studies are based on the Annotated New York Times corpus~\cite{Sandhaus2008}, a newswire corpus with expert-written abstracts.

\subsection{Automatic Corpus Creation}
\noindent\textbf{Event Mention Annotation:} 
Despite many annotation attempts on events~\cite{Pustejovsky2002,Brown2017}, automatic labeling of them in general domain remains an open problem. Most of the previous work follows empirical approaches. For example, \newcite{Chambers2008} consider all verbs together with their subject and object as events. \newcite{Do2011} additionally include nominal predicates, using the nominal form of verbs and lexical items under the \textit{Event} frame in FrameNet~\cite{Baker1998}. 

There are two main challenges in labeling event mentions. First, we need to decide which lexical items are event triggers. Second, we have to disambiguate the word sense to correctly identify events. For example, the word ``phone'' can refer to an entity (a physical phone) or an event (a phone call event). We use FrameNet to solve these problems. We first use a FrameNet based parser: Semafor~\cite{Das2011}, to find and disambiguate triggers into frame classes. We then use the FrameNet ontology to select event mentions.

Our frame based selection method follows the Vendler classes~\cite{Vendler1957}, a four way classification of eventuality: \textit{states}, \textit{activities}, \textit{accomplishments} and \textit{achievements}. The last three classes involve state change, and are normally considered as events. Following this, we create an ``event-evoking frame'' list using the following procedure:

% Specifically, we parse the documents with Semafor and then filter the detected predicates with an ``event-evoking frame'' list, created by the following procedure:
\begin{enumerate}[nolistsep]
\item We keep frames that are subframes of \textit{Event} and \textit{Process} in the FrameNet ontology.
\item We discard frames that are subframes of state, entity and attribute frames, such as  \textit{Entity}, \textit{Attributes}, \textit{Locale}, etc.
\item We manually inspect frames that are not subframes of the above-mentioned ones (around 200) to keep event related ones (including subframes), such as \textit{Arson}, \textit{Delivery}, etc.
\end{enumerate}

This gives us a total of 569 frames. We parse the documents with Semafor and consider predicates that trigger a frame in the list as candidates. We finish the process by removing the light verbs\footnote{Light verbs carry little semantic information: ``appear'', ``be'', ``become'', ``do'', ``have'', ``seem'', ``do'', ``get'', ``give'', ``go'', ``have'', ``keep'', ``make'', ``put'', ``set'', ``take''.} and reporting events\footnote{Reporting verbs are normally associated with the narrator: ``argue'', ``claim'', ``say'', ``suggest'', ``tell''.} from the candidates, similar to previous research~\cite{Recasens2013}. 
% Figure~\ref{fig:annotation_example} shows some example annotations. 
% The full dataset statistics is in Table~\ref{tab:dataset}. 

% \subsection{Salience Labeling from Summary}
% \label{sec:labeling}
\noindent \textbf{Salience Labeling:} For all articles with a human written abstract (around 664,911) in the New York Times Annotated Corpus, we extract event mentions. We then label an event mention as salient if we can find its lemma in the corresponding abstract (\newcite{Mitamura2015} showed that lemma matching is a strong baseline for event coreference.). For example, in Figure~\ref{fig:annotation_example}, event mentions in bold and red are found in the abstract, thus labeled as salient. Data split is detailed in Table~\ref{tab:dataset} and \S\ref{sec:experiments}. 

\input{tables/dataset}

\subsection{Annotation Quality}
While the automatic method enables us to create a dataset at scale, it is important to understand the quality of the dataset. For this purpose, we have conducted two small manual evaluation study.

Our lemma-based salience annotation method is based on the assumption that lemma matching being a strong detector for event coreference. In order to validate this assumption, one of the authors manually examined \(10\) documents and identified \(82\) coreferential event mentions pairs between the text body and the abstract. The automatic lemma rule identifies \(72\) such pairs: \(64\) of these matches human decision, producing a precision of \(88.9\%\) (\(64/72\)) and a recall of \(78\%\) (\(64/82\)). There are \(18\) coreferential pairs missed by the rule.

The next question is: \emph{is an event really important if it is mentioned in the abstract?} Although prior work~\cite{Dunietz2014} shows that the assumption to be valid for entities, we study the case for events. We asked two annotators to manually annotate \(10\) documents (around \(300\) events) using a 5-point Likert scale for salience. We compute the agreement score using Cohen's Kappa~\cite{kappa}. We find the task to be challenging for human: annotators don't agree well on the 5-point scale (Cohen’s Kappa = \(0.29\)). However, if we collapse the scale to binary decisions, the Kappa between the annotators raises to \(0.67\). Further, the Kappa between each annotator and automatic labels are \(0.49\) and \(0.42\) respectively. These agreement scores are also close to those reported in the entity salience tasks~\cite{Dunietz2014}.

While errors exist in the automatic annotation process inevitably, we find the error rate to be reasonable for a large-scale dataset. Further, our study indicates the difficulties for human to rate on a finer scale of salience. We leave the investigation of continuous salience scores to future work.

%% file: tables/dataset.tex
\begin{table}[t]
\centering
\begin{tabular}{l r r r}
\toprule
& Train & Dev & Test \\ \midrule
\# Documents & 526126 & 64000 & 63589  \\\midrule
Avg. \# Word & 794.12 & 790.27 & 798.68\\ \midrule
Avg. \# Events & 61.96 & 60.65 & 61.34\\\midrule
Avg. \# Entities & 197.63 & 196.95 & 198.40\\\midrule
Avg. \# Salience & 8.77 & 8.79 & 8.90\\
\bottomrule
\end{tabular}
\caption{Dataset Statistics.\label{tab:dataset}
}
\end{table}

%% file: 4.feature-based.tex
\section{Feature-Based Event Salience Model}\label{sec:feature_model}
This section presents the feature-based model, including the features and the learning process.

\subsection{Features}
\input{tables/features}
Our features are summarized in Table~\ref{tab:feature}.
 
\noindent\textbf{Basic Discourse Features:} We first use two basic features similar to \newcite{Dunietz2014}: \emph{Frequency} and \emph{Sentence Location}. \emph{Frequency} is the lemma count of the mention's syntactic head word~\cite{Manning2014}. \emph{Sentence Location} is the sentence index of the mention, since the first few sentences are normally more important. These two features are often used to estimate salience~\cite{Barzilay2008,Vossen2015}.

\noindent\textbf{Content Features:} We then design several lexical similarity features, to reflect Grimes' content relatedness~\cite{Grimes1975}. In addition to events, the relations between events and entities are also important. For example, Figure~\ref{fig:annotation_example} shows some related entities in the legal domain, such as ``\textit{prosecutors}'' and ``\textit{court}''. Ideally, they should help promote the salience status for event ``\textit{trial}''. 

Lexical relations can be found both within-sentence (local) or across sentence (global)~\cite{Halliday1976}. We compute the local part by averaging similarity scores from other units in the same sentence. The global part is computed by averaging similarity scores from other units in the document. All similarity scores are computed using cosine similarities on pre-trained embeddings~\cite{word2vec}.  

These lead to 3 content features: \emph{Event Voting}, the average similarity to other events in the document; \emph{Entity Voting}, the average similarity to entities in the document; \emph{Local Entity Voting}, the average similarity to entities in the same sentence. Local event voting is not used since a sentence often contains only 1 event.

\subsection{Model}
A Learning to Rank (\texttt{LeToR}) model~\cite{liu2009learning} is used to combine the features. Let \(ev_i\) denote the \(i\)th event in a document \(d\). Its salience score is computed as:

\begin{align}
\small
f(ev_i, d) = W_f \cdot F(ev_i, d) + b
\end{align}

\noindent where \(F(ev_i, d)\) is the features for \(ev_i\) in \(d\) (Table~\ref{tab:feature}); \(W_f\) and \(b\) are the parameters to learn.

The model is trained with pairwise loss: 
\begin{small}
\begin{align}
\sum_{ev^+, ev^- \in d}   \max(0,  1 - f(ev^+, d) + f(ev^-, d)), \label{eq:salience_loss}\\
\text{w.r.t.}   \text{  } y(ev^+, d) = +1 \text{ }\& \text{ } y(ev^-, d) = -1. \nonumber
\end{align}

\begin{align*}
y(e_i, d) &= \begin{cases}
+1, & \text{if \(e_i\) is a salient entity in } d, \\
-1, & \text{otherwise.}
\end{cases}
\end{align*}
\end{small}

\noindent where \(ev^+\) and \(ev^-\) represent the salient and non-salient events; \(y\) is the gold standard function. Learning can be done by standard gradient methods. 

%% file: tables/features.tex
\begin{table*}[t]
\centering
\begin{tabular}{l l }
\toprule
\textbf{Name} & \textbf{Description} \\ \midrule
\texttt{Frequency} & The frequency of the event lemma in document. \\
\texttt{Sentence Location} & The location of the first sentence that contains the event. \\ \midrule
\texttt{Event Voting} & Average cosine similarity with other events in document. \\ 
\texttt{Entity Voting} & Average cosine similarity with other entities in document. \\
\texttt{Local Entity Voting} & Average cosine similarity with entities in the sentence.\\
\bottomrule
\end{tabular}
\caption{Event Salience Features.\label{tab:feature}
}
\end{table*}

%% file: 5.neural-based.tex
\section{Neural Event Salience Model}\label{sec:neural_model}
% Our feature based method approximates cohesion by lexical similarities. We believe cohesion to be a much richer phenomenon. 
As discussed in \S\ref{sec:introduction}, 
the salience of discourse units is reflected by rich relations beyond lexical similarities, for example, script (``\textit{charge}'' and ``\textit{trial}'') and frame (a ``\textit{trial}'' of ``\textit{attacks}''). The relations between these words are specific to the salience task, thus difficult to be captured by raw cosine scores that are optimized for word similarities. 
In this section, we present a neural model to exploit the embedding space more effectively, in order to capture relations for event salience estimation.

\subsection{Kernel-based Centrality Estimation}
% The raw cosine score between two embedding vectors can be viewed as votes via lexical similarity. 
Inspired by the kernel ranking model~\cite{Xiong2017}, we propose Kernel-based Centrality Estimation (\texttt{KCE}), to find and weight semantic relations of interests, in order to better estimate salience. 

Formally, given a document \(d\), the set of annotated events \(\mathbb{V}=\{ev_1, \ldots ev_i \ldots ,ev_n\}\), \texttt{KCE} first embed an event into vector space:
\(ev_i \xrightarrow{Emb} \overrightarrow{ev_i}\). The embedding function is initialized with pre-trained embeddings. It then extract \(K\) features for each \(ev_i\):

\begin{small}
\begin{align}
\Phi_K(ev_i, \mathbb{V}) &= \{\phi_1(\overrightarrow{ev_i}, \mathbb{V}), \ldots , \\ \nonumber
                        & \quad \quad \phi_k(\overrightarrow{ev_i}, \mathbb{V}), \ldots ,\phi_K(\overrightarrow{ev_i}, \mathbb{V}) \},  \\ 
\phi_k(\overrightarrow{ev_i}, \mathbb{V}) &= \sum_{ev_j \in \mathbb{V}}\exp\left(-\frac{\left(\cos(\overrightarrow{ev_i}, \overrightarrow{ev_j}) - \mu_k\right)^2}{2 \sigma_k^2}\right).
\end{align}
\end{small}

\noindent \(\phi_k(\overrightarrow{ev_i}, \mathbb{V})\) is the \(k\)-th Gaussian kernel with mean \(\mu_k\) and variance \(\sigma_k^2\). It models the interactions between events in its kernel range defined by \(\mu_k\) and \(\sigma_k\).
\(\Phi_K(ev_i, \mathbb{V})\) enforces multi-level interactions among events --- relations that contribute similarly to salience are expected to be grouped into the same kernels. Such interactions greatly improve the capacity of the model with negligible increase in the number of parameters. Empirical evidences~\cite{Xiong2017} have shown that kernels in this form are effective to learn weights for task-specific term pairs.

% Kernels are effective in modeling interactions in Information Retrieval~\cite{Xiong2017}, but have not yet been used in discourse analysis.

The final salience score is computed as:
\begin{align}
\label{eq:event_kernel}
f(ev_i, d) = W_v \cdot \Phi_K(ev_i, \mathbb{V}) + b,
\end{align}
where \(W_v\) is learned to weight the contribution of the certain relations captured by each kernel.

We then use the exact same learning objective as in equation (\ref{eq:salience_loss}). The pairwise loss is first back-propagated through the network to update the kernel weights \(W_v\), assigning higher weights to relevant regions. 
Then the kernels use the gradients to update the embeddings, in order to capture the meaningful discourse relations for salience. 

Since the features and \texttt{KCE} capture different aspects, combining them may give superior performance. This can be done by combining the two vectors in the final linear layer:

\begin{small}
\begin{align}
 f(ev_i, d) &= W_v \cdot \Phi_K(ev_i, \mathbb{V}) + W_f \cdot F(ev_i, d) + b
\end{align}
\end{small}

\vspace{-2em}

\subsection{Integrating Entities into KCE}
\texttt{KCE} is also used to model the relations between events and entities. For example, in Figure~\ref{fig:annotation_example}, the entity ``\textit{court}'' is a frame element of the event ``\textit{trial}''; ``\textit{United States}'' is a frame element of the event ``\textit{war}''. It is not clear which pair contributes more to salience. We again let \texttt{KCE} to learn it. 

Formally, let \(\mathbb{E}\) be the list of entities in the document, i.e. \(\mathbb{E} = \{en_1,\ldots, en_i, \ldots ,en_n\}\), where \(en_i\) is the \(i\)th entity in document \(d\). \texttt{KCE} extracts the kernel features about entity-event relations as follows:

\begin{small}
\begin{align}
\Phi_K(ev_i, \mathbb{E}) &= \{\phi_1(\overrightarrow{ev_i}, \mathbb{E}),\ldots, \\ \nonumber
                         & \quad\quad \phi_k(\overrightarrow{ev_i}, \mathbb{E}), \ldots ,\phi_K(\overrightarrow{ev_i}, \mathbb{E}) \},  \\
\phi_k(\overrightarrow{ev_i}, \mathbb{E}) &= \sum_{en_j \in \mathbb{E}}\exp\left(-\frac{\left(\cos(\overrightarrow{ev_i}, \overrightarrow{en_j}) - \mu_k\right)^2}{2 \sigma_k^2}\right)
\end{align}
\end{small}\\
similarly, \(en_i\) is embedded by: \(en_i \xrightarrow{Emb} \overrightarrow{en_i}\), which is initialized by pre-trained entity embeddings.

We reach the full \texttt{KCE} model by combining all the vectors using a linear layer:

\begin{align}
\nonumber
f(ev_i, d) &= W_e \cdot \Phi_K(ev_i, \mathbb{E}) + W_v \cdot \Phi_K(ev_i, \mathbb{V})\\            &+ W_f \cdot F(ev_i, d) + b
\end{align}
The model is again trained by equation (\ref{eq:salience_loss}).

%% file: 6.experiment.tex
\section{Experimental Methodology}\label{sec:experiments}
This section describes our experiment settings.

\subsection{Event Salience Detection}
\textbf{Dataset:} We conduct our experiments on the salience corpus described in \S\ref{sec:corpus}. Among the 664,911 articles with abstracts, we sample 10\% of the data as the test set and then randomly leave out another 10\% documents for development. Overall, there are 4359 distinct event lexical items, at a similar scale with previous work~\cite{Chambers2008, Do2011}.  
The corpus statistics are summarized in Table~\ref{tab:dataset}.

\noindent\textbf{Input:} The inputs to models are the documents and the extracted events. The models are required to rank the events from the most to least salience.

\noindent\textbf{Baselines:} 
Three methods from previous researches are used as baselines: \emph{Frequency}, \emph{Location} and \emph{PageRank}. The first two are often used to simulate saliency~\cite{Barzilay2008,Vossen2015}. The \emph{Frequency} baseline ranks events based on the count of the headword lemma; the \emph{Location} baseline ranks events using the order of their appearances in discourse. Ties are broken randomly.

Similar to entity salience ranking with PageRank scores~\cite{Xiong2018},  
our \emph{PageRank} baseline runs PageRank on a fully connected graph whose nodes are the events in documents. The edges are weighted by the embedding similarities between event pairs. We conduct supervised PageRank on this graph, using the same pairwise loss setup as in \texttt{KCE}. 
% Similar to \newcite{Xiong2018}, we found it not very effective and the training does not converge well. 
We report the best performance obtained by linearly combining \emph{Frequency} with the scores obtained after a one-step random walk.

\noindent\textbf{Evaluation Metric:}
Since the importance of events is on a continuous scale, the boundary between ``important'' and ``not important'' is vague. 
Hence we evaluate it as a ranking problem. The metrics are the precision and recall value at 1, 5 and 10 respectively. It is adequate to stop at 10 since there are less than 9 salient events per document on average (Table~\ref{tab:dataset}). We also report Area Under Curve (AUC). Statistical significance values are tested by permutation (randomization) test with \(p<0.05\).

\noindent\textbf{Implementation Details:}
We pre-trained word embeddings with 128 dimensions on the whole Annotated New York Times corpus using Word2Vec~\cite{word2vec}. Entities are extracted using the TagMe entity linking toolkit~\cite{Ferragina2010}. Words or entities that appear only once in training are replaced with special ``unknown'' tokens. 

The hyper-parameters of the \texttt{KCE} kernels follow previous literature~\cite{Xiong2017}. There is one exact match kernel (\(\mu=1, \sigma=1e^{-3}\)) and ten soft-match kernels evenly distributed between \((-1, 1)\), i.e. \(\mu \in \{-0.9, -0.7,\ldots ,0.9\}\), with the same \(\sigma=0.1\).

The parameters of the models are optimized by Adam~\cite{Kingma2015}, with batch size 128. The vectors of entities are initialized by the pre-trained embeddings. Event embeddings are initialized by their headword embedding. 

\subsection{The Event Intrusion Test: A Study }\label{sec:extrinsic}
\texttt{KCE} is designed to estimate salience by modeling relations between discourse units. To better understand its behavior, we design the following \textbf{event intrusion test}, following the word intrusion test used to assess topic model quality~\cite{Chang2009}.

\noindent \textbf{Event Intrusion Test:} The test will present to a model a set of events, including: the \textbf{origins}, all events from one document; the \textbf{intruders}, some events from another document. Intuitively, if events inside a document are organized around the core content, a model capturing their relations well should easily identify the intruder(s).

Specifically, we take a bag of unordered events \(\{O_1, O_2, \ldots , O_p\}\), from a document \(O\), as the origins. We insert into it intruders, events drawn from another document, \(I\): \(\{I_1, I_2, \ldots , I_q\}\).
We ask a model to rank the mixed event set \(M = \{O_1, I_1, O_2, I_2, \ldots \}\). We expect a model to rank the intruders \(I_i\) below the origins \(O_i\).

\noindent \textbf{Intrusion Instances:} From the development set, we randomly sample 15,000 origin and intruding document pairs. To simplify the analysis, we only take documents with at least 5 salient events. The intruder events, together with the entities in the same sentences, are added to the origin document. 

\noindent \textbf{Metrics:} AUC is used to quantify ranking quality, where events in \(O\) are positive and events in \(I\) are negative. To observe the ranking among the salient origins, we compute a separate AUC score between the intruders and the salient origins, denoted as SA-AUC\@. In other words, SA-AUC is the AUC score on the list with non-salient origins removed.

\noindent \textbf{Experiments Details:} We take the full \texttt{KCE} model to compute salient scores for events in the mixed event set \(M\), which are directly used for ranking. Frequency is recounted. All other features (Table~\ref{tab:feature}) are set to 0 to emphasize the relational aspects, 
% If \texttt{KCE} can rank the intruders low, it will obtain high AUC. 

We experiment with two settings: 1.\ adding only the salient intruders. 2.\ adding only the non-salient intruders. Under both settings, the intruders are added one by one, allowing us to observe the score change regarding the number of intruders added. For comparison, we add a \emph{Frequency} baseline, that directly ranks events by the Frequency feature.

%% file: 7.evaluation.tex
\section{Evaluation Results}
This section presents the evaluations and analyses.

\subsection{Event Salience Performance}\label{sec:results}
\input{tables/results}
\input{tables/ablation}

We summarize the main results in Table~\ref{tab:main_results}. 

\noindent\textbf{Baselines:} \emph{Frequency} is the best performing baseline. Its precision at 1 and 5 are higher than \(40\%\). \emph{PageRank} performs worse than \emph{Frequency} on all the precision and recall metrics. \emph{Location} performs the worst.

\noindent\textbf{Feature Based:} \texttt{LeToR} outperforms the baselines significantly on all metrics. Particularly, its P@1 value outperforms the \emph{Frequency} baseline the most (\(4.64\%\)), indicating a much better estimation on the most salient event. In terms of AUC, \texttt{LeToR} outperforms \emph{Frequency} by a large margin (\(11.19\%\) relative gain). 

\noindent\textbf{Feature Ablation:} To understand the contribution of individual features, we conduct an ablation study of various feature settings in Table~\ref{tab:ablation}. We gradually add feature groups to the \emph{Frequency} baseline.
The combination of \emph{Location} (sentence location) and \emph{Frequency} almost sets the performance for the whole model. Adding each voting feature individually produces mixed results. However, adding all voting features improves all metrics. Though the margin is small, 4 of them are statistically significant over \emph{Frequency+Location}.

\noindent \textbf{Kernel Centrality Estimation:} The \texttt{KCE} model further beats \texttt{LeToR} significantly on all metrics, by around \(5\%\) on AUC and precision values, and by around \(10\%\) on the recall values. Notably, the P@1 score is much higher, reaching \(50\%\). The large relative gain on all the recall metrics and the high performance on precision show that \texttt{KCE} works really well on the top of the rank list.

\noindent \textbf{Kernel Ablation:} To understand the source of performance gain of \texttt{KCE}, we conduct an ablation study by removing its components: \texttt{-E} removes of entity kernels; \texttt{-EF} removes the entity kernels and the features. We observe a performance drop in both cases. Without entities and features, the model only using event information still performs similarly to \emph{Frequency}.
% (worse on 3 metrics, but better on 3 others). 
The drops are also a reflection of the small number of events (\(\approx\) 60 per document) comparing to entities (\(\approx\) 200 per document).
% : removing the entity information greatly reduce the amount of interactions. 
The study indicates that the relational signals and features contain different but both important information.

\begin{figure}[ht]
\includegraphics[width=\linewidth]{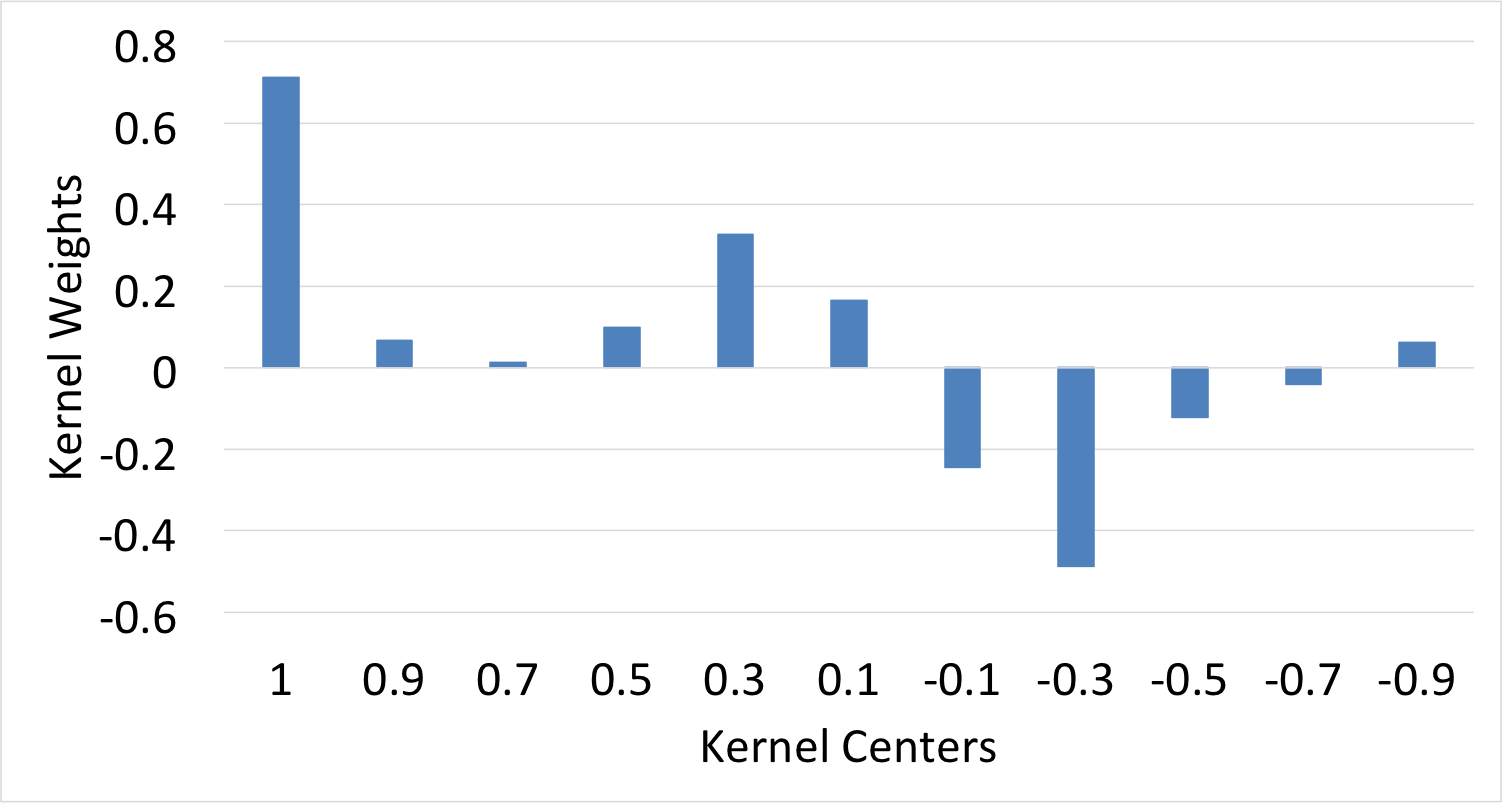}
\caption{Learned Kernel Weights of \texttt{KCE}}\label{fig:kernel_weights}
\end{figure}

\noindent \textbf{Discussion:} The superior results of \texttt{KCE} demonstrate its effectiveness in predicting salience. So what additional information does it capture?
We revisit the changes made by \texttt{KCE}: 1.\ it adjusts the embeddings during training. 2.\ it introduces weighted soft count kernels. 
However, the \emph{PageRank} baseline also does embedding tuning but produces poor results, thus the second change should be crucial. We plot the learned kernel weights of \texttt{KCE} in Figure~\ref{fig:kernel_weights}. Surprisingly, the salient decisions are not linearly related, nor even positively correlated to the weights. In fact, besides the ``Exact Match'' bin, the highest absolute weights actually appear at 0.3 and -0.3. 
This implies that embedding similarities do not directly imply salience, breaking some assumptions of the feature based model and \emph{PageRank}.

\input{tables/kernel_sim_examples}

\begin{figure*}[ht]
\includegraphics[width=0.49\linewidth]{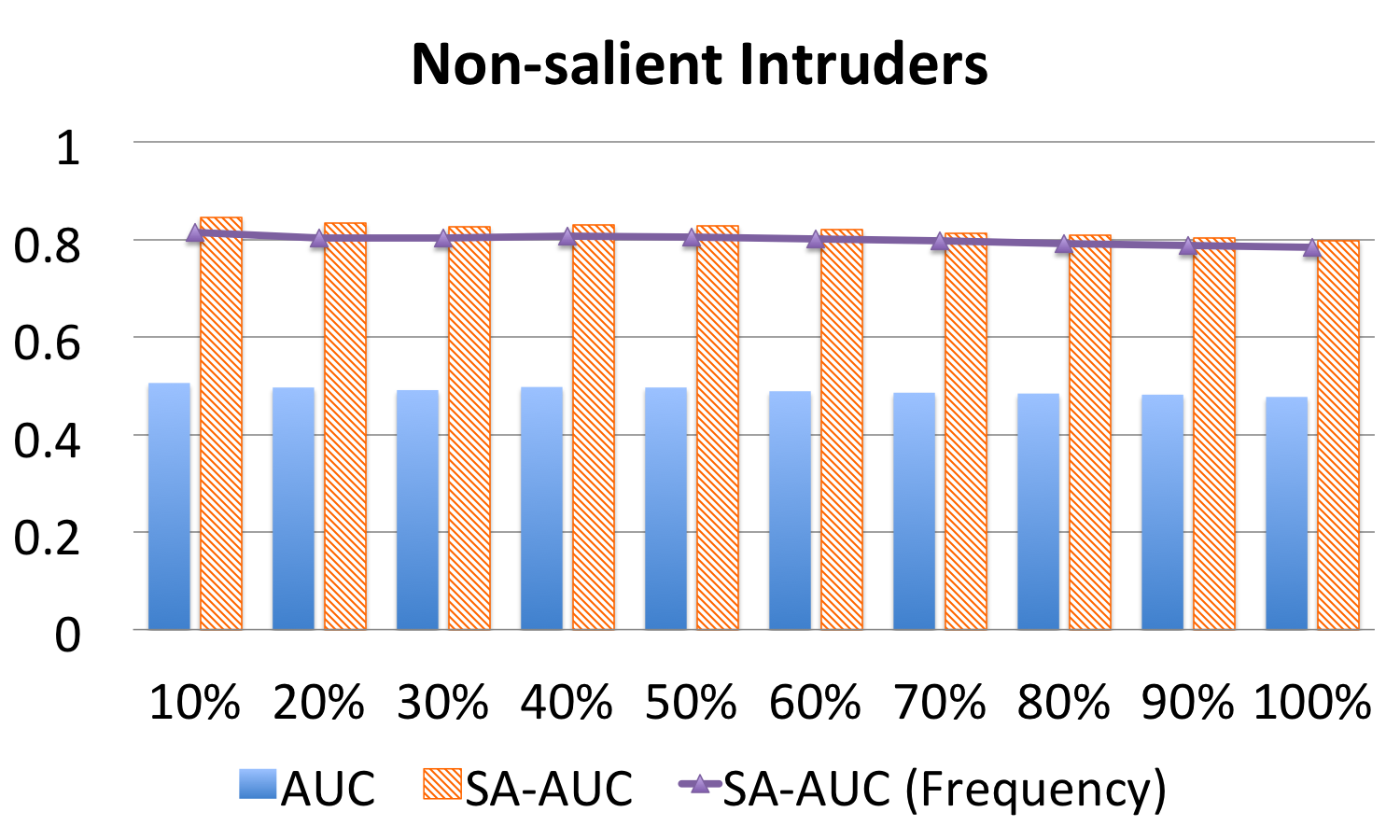}
\includegraphics[width=0.49\linewidth]{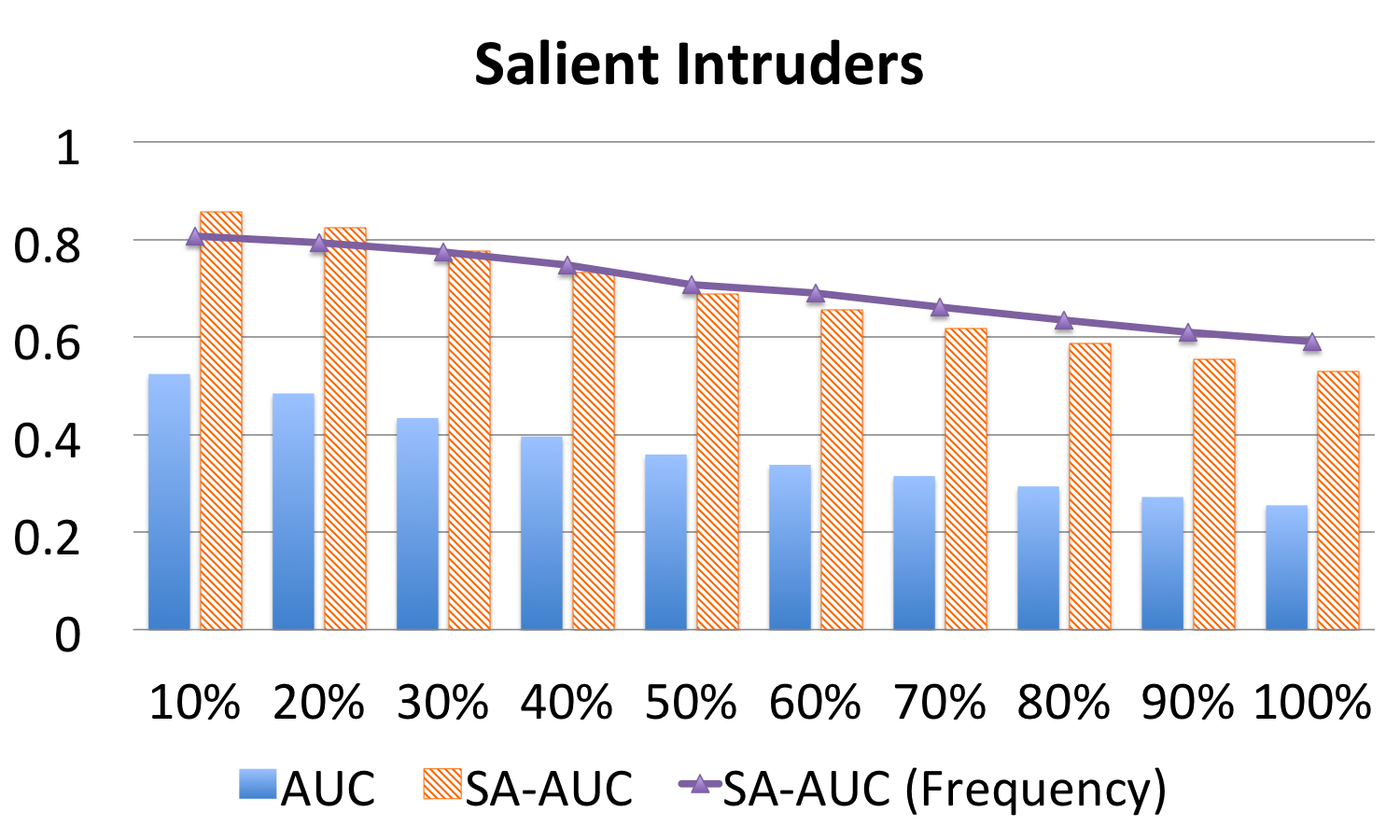}
\caption{Intruder study results. X-axis shows the percentage of intruders inserted. Y-axis is the AUC score scale. The left and right figures are results from salient and non-salient intruders respectively. The blue bar is AUC\@. The orange shaded bar is SA-AUC\@. The line shows the SA-AUC of the frequency baseline.}\label{fig:intruder}
\end{figure*}

\noindent\textbf{Case Study:} We inspect some pairs of events and entities in different kernels and list some examples in Table~\ref{tab:kernel_examples}. The pre-trained embeddings are changed a lot. Pairs of units with different raw similarity values are now placed in the same bin. The pairs in Table 3 exhibit interesting types of relations: e.g.,``\textit{arrest-charge}'' and ``\textit{attack-kill}'' form script-like chains; ``911 attack'' forms a quasi-identity relation~\cite{Recasens2010} with ``attack''; ``business'' and ``increase'' are candidates as frame-argument structure. While these pairs have different raw cosine similarities, they are all useful in predicting salience. \texttt{KCE} learns to gather these relations into bins assigned with higher weights, which is not achieved by pure embedding based methods. The \texttt{KCE} has changed the embedding space and the scoring functions significantly from the original space after training. This partially explains why the raw voting features and PageRank are not as effective. 

\subsection{Intrusion Test Results}\label{sec:intrusion_eval}

Figure~\ref{fig:intruder} plots results of the intrusion test
% , showing AUC with different percentages of the intruders inserted (X-axis)
. 
The left figure shows the results of setting 1: adding non-salient intruders. The right one shows the results of setting 2: adding salient intruders. The AUC is 0.493 and the SA-AUC is 0.753 if all intruders are added.

The left figure shows that \texttt{KCE}  successfully finds the non-salient intruders.  The SA-AUC is higher than 0.8. Yet the AUC scores, which include the rankings of non-salience events, are rather close to random. This shows that the salient events in the origin documents form a more cohesive group, making them more robust against the intruders; the non-salient ones are not as cohesive.
% The non-salient events can be understood as background events, similar to prior work~\cite{Cheung2013}.

In both settings, \texttt{KCE} produces higher SA-AUC than \emph{Frequency} at the first 30\%. However, in setting 2, \texttt{KCE} starts to produce lower SA-AUC than  \emph{Frequency} after 30\%, then gradually drops to 0.5 (random). This phenomenon is expected since the asymmetry between origins and intruders allow \texttt{KCE} to distinguish them at the beginning.  When all intruders are added, \texttt{KCE} performs worse because it relies heavily on the relations, which can be also formed by the salient intruders. This phenomenon is observed only on the salient intruders, which again confirms the cohesive relations are found among salient events.
% , not on the non-salient ones. It is interesting considering that salient events account for only 14\% of all events on average

% (Table~\ref{tab:dataset}).

% We conclude that \texttt{KCE} has certain ability to detect intruders. In addition, 
In conclusion, we observe that the salient events form tight groups connected by discourse relations while the non-salient events are not as related. The observations imply that the main scripts in documents are mostly anchored by small groups of salient events (such as the ``Trial'' script in Example~\ref{fig:annotation_example}). Other events may serve as ``backgrounds''~\cite{Cheung2013}.
Similarly, \newcite{Choubey2018} find that relations like event coreference and sequence are important for saliency. 

%% file: tables/results.tex
\begin{table*}[t]
\small
\centering
\begin{tabular}{l | l r | l r | l r | l r }
\toprule
\bf{Method}& \multicolumn{2}{c}{\bf{P@01}}& \multicolumn{2}{c}{\bf{P@05}}& \multicolumn{2}{c}{\bf{P@10}} & \multicolumn{2}{c}{\bf{AUC}}\\ \midrule

\texttt{Location}
 & 0.3555 & -- & 0.3077 & -- & 0.2505 & -- & 0.5226 & --\\
 
\texttt{PageRank}
 & 0.3628 & -- & 0.3438 & -- & 0.3007 & -- & 0.5866 & --\\

\texttt{Frequency}
 & \(0.4542\) & --  & \(0.4024\) & --  & \(0.3445\) & -- & \(0.5732\) & -- \\
 
 \midrule

\texttt{LeToR}
 & \({0.4753}^{\dagger }\) & \( +4.64\%  \)

 & \({0.4099}^{\dagger }\) & \( +1.87\%  \)

 & \({0.3517}^{\dagger }\) & \( +2.10\%  \)

 & \({0.6373}^{\dagger }\) & \( +11.19\%  \)

\\ 
 
\texttt{KCE (-EF)}
 & \({0.4420}\) & \( -2.69\%  \)

 & \({0.4038}\) & \( +0.34\%  \)

 & \({0.3464}^{\dagger }\) & \( +0.54\%  \)

 & \({0.6089}^{\dagger }\) & \( +6.23\%  \)

\\
 
 \texttt{KCE (-E)}
 & \({0.4861}^{\dagger \ddagger }\) & \( +7.01\%  \)

 & \({0.4227}^{\dagger \ddagger }\) & \( +5.04\%  \)

 & \({0.3603}^{\dagger \ddagger }\) & \( +4.58\%  \)

 & \({0.6541}^{\dagger \ddagger }\) & \( +14.12\%  \)

\\

 \texttt{KCE}
 & \({0.5049}^{\dagger \ddagger }\) & \( +11.14\%  \)

 & \({0.4277}^{\dagger \ddagger }\) & \( +6.29\%  \)

 & \({0.3638}^{\dagger \ddagger }\) & \( +5.61\%  \)

 & \({0.6557}^{\dagger \ddagger }\) & \( +14.41\%  \)

\\\midrule\midrule

\bf{Method}& \multicolumn{2}{c}{\bf{R@01}}& \multicolumn{2}{c}{\bf{R@05}}& \multicolumn{2}{c}{\bf{R@10}} & \multicolumn{2}{c}{\bf{W/T/L}}\\ \midrule

\texttt{Location}
 & \(0.0807\) & -- & \(0.2671\) & -- & \(0.3792\) & -- & \multicolumn{2}{c}{--/--/--}\\

\texttt{PageRank}
 & \(0.0758\) & -- & \(0.2760\) & -- & \(0.4163\)  & -- & \multicolumn{2}{c}{--/--/--}\\

\texttt{Frequency}
 & \(0.0792\) & --  & \(0.2846\) & --  & \(0.4270\) & --  & \multicolumn{2}{c}{--/--/--}\\
 
 \midrule
  
\texttt{LeToR}
 & \({0.0836}^{\dagger }\) & \( +5.61\%  \)

 & \({0.2980}^{\dagger }\) & \( +4.70\%  \)

 & \({0.4454}^{\dagger }\) & \( +4.31\%  \)

& \multicolumn{2}{c}{8037 / 48493 / 6770}

\\ 
 
\texttt{KCE (-EF)}

 & \({0.0714}\) & \( -9.77\%  \)

 & \({0.2812}\) & \( -1.18\%  \)

 & \({0.4321}^{\dagger }\) & \( +1.20\%  \)

& \multicolumn{2}{c}{6936 / 48811 / 7553}

\\

 \texttt{KCE (-E)}
 & \({0.0925}^{\dagger \ddagger }\) & \( +16.78\%  \)

 & \({0.3172}^{\dagger \ddagger }\) & \( +11.46\%  \)

 & \({0.4672}^{\dagger \ddagger }\) & \( +9.41\%  \)

& \multicolumn{2}{c}{11676 / 43294 / 8330}

\\
 
 \texttt{KCE}
 & \({0.0946}^{\dagger \ddagger }\) & \( +19.44\%  \)

 & \({0.3215}^{\dagger \ddagger }\) & \( +12.96\%  \)

 & \({0.4719}^{\dagger \ddagger }\) & \( +10.51\%  \)

& \multicolumn{2}{c}{12554 / 41461 / 9285}

\\
\bottomrule
\end{tabular}
\caption{Event salience performance. (-E) and (-F) marks removing Entity information and Features from the full KCM model. The relative performance differences are computed against \texttt{Frequency}. W/T/L are the number of documents a method wins, ties, and loses compared to \texttt{Frequency}. \(\dagger\) and \(\ddagger\) mark the statistically significant improvements over \texttt{Frequency}\(^\dagger\), \texttt{LeToR}\(^\ddagger\) respectively.\label{tab:main_results}
}
\end{table*}

%% file: tables/ablation.tex
\begin{table*}[t]
\centering
\small
\begin{tabular}{l l l l l l l l}
\toprule
\textbf{Feature Groups} & P@1 & P@5 & P@10 & R@1 & R@5 & R@10 & AUC
 \\ \midrule
\texttt{Loc} & \(0.3548\) & \(0.3069\) & \(0.2497$  & \(0.0807\) & \(0.2671\) & \(0.3792\) & \(0.5226\) \\ 
\texttt{Frequency} & \(0.4536\) & \(0.4018\) & \(0.3440$  & \(0.0792\) & \(0.2846\) & \(0.4270\) & \(0.5732\) \\
 \midrule
+ \texttt{Loc} & \(0.4734\) & \(0.4097\) & \(0.3513\) & \(0.0835\) & \(0.2976\) & \(0.4436\) & \(0.6354\)\\ 
+ \texttt{Loc} + \texttt{Event} & \(0.4726\) & \(0.4101^{\dagger}\) & \(0.3516\) & \(0.0831\) & \(0.2969\) & \(0.4431\) & \(0.6365^{\dagger}$\\  
+ \texttt{Loc} + \texttt{Entity} & \(0.4739\) & \(0.4100\) & \(0.3518\) & \(0.0812\) & \(0.2955\) & \(0.4418\) & \(0.6374$\\  
+ \texttt{Loc} + \texttt{Entity} + \texttt{Event} & \(0.4739\) & \(0.4100\) & \(0.3518^{\dagger}\) & \(0.0832\) & \(0.2974\) & \(0.4452^{\dagger}\) & \(0.6374^{\dagger}$\\ 
+ \texttt{Loc} + \texttt{Entity} + \texttt{Event} + \texttt{Local} & \({0.4754}^{\dagger}\) & \(0.4100\) & \({0.3517}^{\dagger}\) & \({0.0837}\) & \({0.2981}\) & \({0.4454}^{\dagger}\) & \({0.6373}^{\dagger}$\\ 
% \hline
\bottomrule
\end{tabular}
\caption{Feature Ablation Results. \texttt{+} sign indicates the additional features to \texttt{Frequency}. \texttt{Loc} is the sentence location feature. \texttt{Event} is the event voting feature. \texttt{Entity} is the entity voting feature. \texttt{Local} is the local entity voting feature. \({\dagger}\) marks the statistically significant improvements over + \texttt{Loc}.\label{tab:ablation}
}
\end{table*}

%% file: tables/kernel_sim_examples.tex
\begin{table}[t]
\centering
\begin{tabular}{l l r r}
\toprule
 & & \textbf{Word2Vec} & \textbf{KCE}\\ \midrule
attack & kill & 0.69 & 0.3\\
arrest & charge & 0.53 & 0.3 \\
USA (E) & war & 0.46 & 0.3\\
911 attack (E) & attack & 0.72 &0.3 \\
attack & trade & 0.42 & 0.9 \\
hotel (E)  & travel & 0.49 & 0.9\\
charge & murder & 0.49 & 0.7 \\
business (E) & increase & 0.43 & 0.7\\
attack & walk & 0.44 & -0.3 \\
people (E) & work & 0.40 & -0.3\\
\bottomrule
\end{tabular}
\caption{Similarities between event entity pairs. \textbf{Word2vec} shows the cosine similarity in pre-trained embeddings. \textbf{KCE} lists their closest kernel mean after training. (E) marks entities.\label{tab:kernel_examples}
}
\end{table}

%% file: 8.conclusion.tex
\section{Conclusion}
We propose two salient detection models, based on lexical relatedness and semantic relations. The feature-based model with lexical similarities is effective, but cannot capture semantic relations like scripts and frames. The \texttt{KCE} model uses kernels and embeddings to capture these relations, thus outperforms the baselines and feature-based models significantly. All the results are tested on our newly created large-scale event salience dataset.  While the automatic method inevitably introduces noises to the dataset, the scale enables us to study complex event interactions, which is infeasible via costly expert labeling. 

Our case study shows that the salience model finds and utilize a variety of discourse relations: script chain (\textit{attack} and \textit{kill}), frame argument relation (\textit{business} and \textit{increase}), quasi-identity (\textit{911 attack} and \textit{attack}). Such complex relations are not as prominent in the raw word embedding space. The core message is that a salience detection module automatically discovers connections between salience and relations. This goes beyond prior centering analysis work that focuses on lexical and syntax and provide a new semantic view from the script and frame perspective.

In the intrusion test, we observe that the small number of salient events are forming tight connected groups. While \texttt{KCE} captures these relations quite effectively, it can be confused by salient intrusion events. The phenomenon indicates that the salient events are tightly connected, which form the main scripts of documents.
% while the rest serve as the background.
% which may form the background of texts.

% Estimating salience of discourse units is important for natural language understanding.
This paper empirically reveals many interesting connections between discourse phenomena and salience. The results also suggest that core script information may reside mostly in the salient events. Limited by the data acquisition method, this paper only models discourse salience as binary decisions. However, salience value may be continuous and may even have more than one aspects. 
In the future, we plan to investigate these complex settings. Another direction of study is large-scale semantic relation discovery, for example, frames and scripts, with a focus on salient discourse units.